\documentclass[11pt,a4paper]{article}

\usepackage[hyperref]{acl2020}
\usepackage{times}
\usepackage{latexsym}

\usepackage{amsmath}

\usepackage{graphicx}
\usepackage{adjustbox}

\usepackage{amssymb}
\usepackage{bbold}
\usepackage{subfigure}
\usepackage{algorithm}
\usepackage[noend]{algpseudocode}
\usepackage[toc,page]{appendix}
\usepackage{url}            
\usepackage{booktabs}       
\usepackage{amsfonts}       
\usepackage{nicefrac}       
\usepackage{microtype}      

\usepackage{paralist}

\usepackage[capitalise]{cleveref}

\usepackage{enumitem}

\usepackage{xcolor}
\usepackage{paralist}

\newcommand{\rvj}{\textsc{RevExpl}}
\newcommand{\attentionesnli}{\textsc{ExplainThenPredictAttention}}
\newcommand{\attention}{\textsc{EtPA}}

\newcommand{\bfx}{\mathbf{x}}
\newcommand{\bfr}{\mathbf{\hat{x}}}
\newcommand{\bfe}{\mathbf{e}}

\newcommand{\bfi}{\mathbf{\hat{e}}}

\newcommand{\bfxd}{\mathbf{x}_{c}}
\newcommand{\bfxi}{\mathbf{x}_{v}}
\newcommand{\bfri}{\mathbf{\hat{x}}_{v}}

\newcommand{\ie}{i.e.,}

\usepackage{microtype}

\aclfinalcopy 

\title{Make Up Your Mind! Adversarial Generation of \\ Inconsistent Natural Language Explanations}

\author{Oana-Maria Camburu$^{1,4}$ \quad Brendan Shillingford$^{1,2}$ \quad Pasquale Minervini$^{3}$ \\
\bf \large Thomas Lukasiewicz$^{1,4}$ \quad Phil Blunsom$^{1,2}$ \\
$^{1}$ University of Oxford \quad $^{2}$ DeepMind, London \\
$^{3}$ University College London \quad $^{4}$ Alan Turing Institute, London \\
\texttt{firstname.lastname@cs.ox.ac.uk} \quad \texttt{p.minervini@ucl.ac.uk}
}

\date{}

\begin{document}

\maketitle

\begin{abstract}
%
%
To increase trust in artificial intelligence systems, a promising research direction consists of designing neural models capable of generating natural language explanations for their predictions.
In this work, we show that such models are nonetheless prone to generating mutually inconsistent explanations, such as ``\emph{Because there is a dog in the image.}'' and ``\emph{Because there is no dog in the} [same] \emph{image.}'', 
exposing flaws in either the decision-making process of the model or in the generation of the explanations. 
We introduce a simple yet effective adversarial framework for sanity checking models against the generation of inconsistent natural language explanations.
Moreover, as part of the framework, we address the problem of adversarial attacks with \emph{full} target sequences, a scenario that was not previously addressed in sequence-to-sequence attacks. 
Finally, we apply our framework on a state-of-the-art neural natural language inference model that provides natural language explanations for its predictions. 
Our framework shows that this model is capable of generating a significant number of inconsistent explanations.
\end{abstract}

\begin{table*}[t]
\resizebox{\textwidth}{!}{%
\begin{tabular}{p{12cm}|p{12cm}}

\toprule

\multicolumn{2}{c}{\begin{tabular}[c]{@{}c@{}} \textsc{Premise:} A guy in a red jacket is snowboarding in midair.\end{tabular}}

\\

\begin{tabular}[c]{@{}l@{}}
\textsc{Original Hypothesis:} A guy is outside in the snow. \\
\textsc{Predicted Label:} entailment \\
\textsc{Original explanation:} {\bf Snowboarding is done outside.}\end{tabular} &

\begin{tabular}[c]{@{}l@{}}
\textsc{Reverse Hypothesis:} The guy is outside. \\
\textsc{Predicted label:} contradiction \\
\textsc{Reverse explanation:} {\bf Snowboarding is not done outside.} \end{tabular}

\\

\midrule

\multicolumn{2}{c}{\begin{tabular}[c]{@{}c@{}} \textsc{Premise:} A man talks to two guards as he holds a drink.\end{tabular}} 

\\

\begin{tabular}[c]{@{}l@{}}
\textsc{Original Hypothesis:} The prisoner is talking to two guards in the \\ prison cafeteria. \\
\textsc{Predicted Label:} neutral \\
\textsc{Original explanation:} {\bf The man is not necessarily a prisoner.} \end{tabular}
& 
\begin{tabular}[c]{@{}l@{}}
\textsc{Reverse Hypothesis:} A prisoner talks to two guards. \\
\textsc{Predicted Label:} entailment \\
\textsc{Reverse explanation:} {\bf A man is a prisoner.}
\\
\end{tabular} 

\\

\midrule

\multicolumn{2}{c}{\begin{tabular}[c]{@{}c@{}}
\textsc{Premise:} Two women and a man are sitting down eating and drinking various items.\end{tabular}}

\\

\begin{tabular}[c]{@{}l@{}}
\textsc{Original Hypothesis:} Three women are shopping at the mall. \\
\textsc{Predicted label:} contradiction \\
\textsc{Original explanation:} \textbf{There are either two women and a man or} \\ \textbf{three women.} \\
\end{tabular}                                
& 
\begin{tabular}[c]{@{}l@{}}
\textsc{Reverse Hypothesis:} Three women are sitting down eating. \\
\textsc{Predicted label:} neutral \\
\textsc{Reverse explanation:} \textbf{Two women and a man are three women.} \\
\end{tabular}

\\

\bottomrule

\end{tabular}%
}
\caption{Examples of detected inconsistent explanations -- the reverse hypotheses generated by our method (right) are realistic.} 
\label{inconsist-examples}
\end{table*}

\section{Introduction}\label{intro}
In order to explain the predictions produced by accurate yet black-box neural models, a growing number of works propose extending these models with natural language explanation generation modules, thus obtaining models that explain themselves in human language~\citep{DBLP:conf/eccv/HendricksARDSD16,esnli,zeynep,cars, math}.

In this work, we first draw attention to the fact that such models, while appealing, are nonetheless prone to generating inconsistent explanations. We define two explanations to be inconsistent if they provide 
contradictory arguments about the instances and predictions that they aim to explain. For example, consider a visual question answering (VQA) task~\citep{zeynep} and two instances where the image is the same but the questions are different, say ``\emph{Is there an animal in the image?}'' and ``\textit{Can you see a Husky in the image?}''. If for the first instance a model predicts ``\emph{Yes.}'' and generates the explanation ``\emph{Because there is a dog in the image.}'', while for the second instance the \emph{same} model predicts ``\emph{No.}'' and generates the explanation ``\textit{Because there is no dog in the image.}'', then the model is producing inconsistent explanations. 
%

Inconsistent explanations reveal at least one of the following undesired 
behaviors:
\begin{inparaenum}[\itshape (i\upshape)]
\item at least one of the explanations is not faithfully describing the decision mechanism of the model, or
\item the model relied on a faulty decision mechanism for at least one of the instances.
\end{inparaenum}
Note that, for a pair of inconsistent explanations, further investigation would be needed to conclude which of these two 
behaviors 
is the actual one (and might vary for each instance). Indeed, a pair of inconsistent explanations does not necessarily imply at least one unfaithful explanation. In our previous example, if the image contains a dog, it is possible that the model identifies the dog when it processes the image together with the first question, and that the model does not identify the dog when it processes the image together with the second question, hence both explanations would faithfully reflect the decision mechanism of the model even if they are inconsistent. Similarly, a pair of inconsistent explanations does not necessarily imply that the model relies on a faulty decision mechanism, because the explanations may not faithfully describe the decision mechanism of the model. We here will not investigate the problem of identifying which of the two undesired behaviors is true for a pair of inconsistent explanations.  
%
%
%
%

In this work, we introduce a framework for checking if models are robust against generating inconsistent natural language explanations. 
Given a model $m$ that produces natural language explanations for its predictions, and an instance $\bfx$, our framework aims to generate inputs $\bfr$ that cause the model to produce explanations that are inconsistent with the explanation produced for $\bfx$. Thus, our framework falls under the category of \emph{adversarial methods}, i.e., searching for inputs that cause a model to produce undesired answers~\citep{DBLP:conf/pkdd/BiggioCMNSLGR13, DBLP:journals/corr/SzegedyZSBEGF13}. 

As part of our framework, we address the problem of adversarial attacks with \emph{full} target sequences, a scenario that has not been previously addressed in sequence-to-sequence attacks, and which can be useful for other areas, such as dialog systems.
Finally, we apply our framework on a state-of-the-art neural natural language inference model that generates natural language explanations for its decisions~\citep{esnli}. We show that this model can generate a significant number of inconsistent explanations.
\section{Method}
Given a model $m$ that can jointly produce predictions and natural language explanations, we propose a framework that, for any given instance $\bfx$, attempts to generate new instances for which the model produces explanations that are inconsistent with the explanation produced for $\bfx$; we refer to the latter as $\bfe_{m}(\bfx)$.
%


%
We approach the problem in two high-level steps. Given an instance $\bfx$, 
\begin{inparaenum}[\itshape (A\upshape)]
\item we create a list of explanations that are inconsistent with the explanation generated by the model on $\bfx$, and \label{A}
\item given an inconsistent explanation from the list created in \ref{A}, we find an input that causes the model to generate this precise inconsistent explanation. \label{B}
\end{inparaenum}

\paragraph{Setup.} Our setup has three desired properties that make it different from commonly researched adversarial settings in natural language processing:
\begin{itemize}
\item At step (\ref{B}), the model has to generate a \emph{full target sequence}: the goal is to generate the \emph{exact} explanation that was identified at step (\ref{A}) as inconsistent with the explanation $\bfe_{m}(\bfx)$.
%
%
%
\item Adversarial inputs do not have to be a paraphrase or a small perturbation of the original input, since our objective is to generate inconsistent explanations rather than incorrect predictions --- these can eventually happen as a byproduct.
\item Adversarial inputs have to be realistic to the task at hand. 
\end{itemize}

To our knowledge, this work is the first to tackle this problem setting, especially due to the challenging requirement of generating a full target sequence --- see \cref{related} for comparison with existing works.
%
%


%
%

%
%
%

%
\paragraph{Context-dependent inconsistencies.}
%
In certain tasks, instances consist of a context (such as an image or a paragraph), and some assessment to be made about the context (such as a question or a hypothesis).
Since explanations may refer (sometimes implicitly) to the context, the assessment of whether two explanations are inconsistent may also depend on it. For example, in VQA, the inconsistency of the two explanations ``\emph{Because there is a dog in the image.}'' and ``\emph{Because there is no dog in the image.}'' depends on the image. However, if the image is the same, the two explanations are inconsistent regardless of which questions were asked on that image. 

For such a reason, given an instance $\bfx$, we differentiate between parts of the instance that will remain fixed in our method (referred to as \emph{context parts} and denoted as $\bfxd$) and parts of the instance that our method will vary in order to obtain inconsistencies (referred to as \emph{variable parts} and denoted as $\bfxi$). Hence, $\bfx = (\bfxd, \bfxi)$. In our VQA example, $\bfxd$ is the image, and $\bfxi$ is the question.
%
%
%

\paragraph{Stand-alone inconsistencies.}
Furthermore, we note that there are cases for which explanations are inconsistent regardless of the input. For example, explanations formed purely of background knowledge such as ``\emph{A woman is a person.}'' and ``\emph{A woman is not a person.}''\footnote{Which was generated by the model in our experiments.} are always inconsistent (and sometimes outrageous), regardless of the instances that lead to them. 
For these cases, our method can treat the whole input as variable, i.e., $\bfxd = \emptyset$ and $\bfri = \bfx$.

\paragraph{Steps.}\label{steps}
Our adversarial framework consists of the following steps:
\begin{enumerate}[wide, labelwidth=!, labelindent=0pt]
    \item Reverse the explanation generator module of model $m$ by training a \rvj~model to map from the generated explanation and the context part of the input to the variable part of the input, \ie{} $\rvj(\bfxd, \bfe_{m}(\bfx)) = \bfxi$.
    \item \label{s2} For each explanation $\bfe = \bfe_m(\bfx)$:
    \begin{enumerate}
        \item \label{rules} Create a list of statements that are inconsistent with $\bfe$, we call it $\mathcal{I}_{\bfe}$. 
        \item \label{b} Query \rvj~on each $\bfi \in \mathcal{I}_{\bfe}$ and the context $\bfxd$. Get the new variable part $\bfri = \rvj(\bfxd, \bfi)$ of a \emph{reverse input} $\bfr = (\bfxd, \bfri)$, which \emph{may} cause the $m$ to produce inconsistent explanations.
        \item Query $m$ on each reverse input to get a \emph{reverse explanation} $\bfe_{m}(\bfr)$.
        \item \label{check_rules} Check if each reverse explanation $\bfe_{m}(\bfr)$ is indeed inconsistent with $\bfe$ by checking if $\bfe_{m}(\bfr) \in \mathcal{I}_{\bfe}$. 
    \end{enumerate}
    
\end{enumerate}
To execute step (\ref{rules}), note that explanations are by nature logical sentences. Hence, for any task, one may define a set of logical rules to transform an explanation into an inconsistent counterpart, such as negation or replacement of task-essential tokens with task-specific antonyms. For example, in explanations for self-driving cars~\citep{cars}, one can replace ``\emph{green light}'' with ``\emph{red light}'', or ``\emph{the road is empty}'' with ``\emph{the road is crowded}'' (which are task-specific antonyms), to get inconsistent (and hazardous) explanations such as ``\emph{The car accelerates because there is a red light.}''.
Another strategy to obtain inconsistent explanations consists of swapping explanations from mutually exclusive labels. For example, assume a recommender system predicts that movie \emph{X} is a bad recommendation for user \emph{Y} ``\emph{because X is a horror movie.}'', implying that user Y does not like horror movies. If it also predicts that movie \emph{Z} is a good recommendation to the same user \emph{Y} ``\emph{because Z is a horror movie.}'', then we have an inconsistency, as the latter would imply that user Y likes horror movies.

While this step requires a degree of specific adjustment to the task at hand, we consider it a small price to pay to ensure that the deployed system is coherent. Also, note that this step can eventually be automated, for example, by training a neural network to generate task-specific inconsistencies after crowd-sourcing a dataset of inconsistent explanations for a task at hand --- we leave this as future work.
Finally, to execute step (\ref{check_rules}), our framework currently checks for an exact string match between a reverse explanation and any of the inconsistent explanations created at step (\ref{rules}). Alternatively, one can train a model to identify if a pair of explanations forms an inconsistency, which we also leave as future work.
\section{Experiments}\label{experiments}
We consider the task of natural language inference (NLI)~\citep{snli}, which consists of detecting whether a pair of sentences, called \emph{premise} and \emph{hypothesis}, are in a relation of: \emph{entailment}, if the premise entails the hypothesis; \emph{contradiction}, if the premise contradicts the hypothesis; or \emph{neutral}, if neither entailment nor contradiction holds.
For example,  a pair with premise ``\emph{Two doctors perform surgery on patient.}'' and hypothesis ``\emph{Two doctors are performing surgery on a man.}'' constitutes a neutral pair.
%

%
The SNLI corpus~\citep{snli} of $\sim\!\!570$K such human-written instances enabled a plethora of works on this task~\citep{RocktaschelGHKB15,NSE,intra-attention-LIU}.
Recently, \citet{esnli} augmented SNLI with crowd-sourced free-form explanations of the ground-truth label, called e-SNLI. An explanation from e-SNLI for the neutral pair above is ``\emph{Not every patient is a man.}''. 
Their best model for generating explanations, called \attentionesnli~(hereafter called \attention), is a sequence-to-sequence attention model that uses two bidirectional LSTM networks~\citep{lstm} for encoding the premise and hypothesis, and an LSTM decoder for generating the explanation while separately attending over the tokens of the premise and hypothesis. Subsequently, they predict the label solely based on the explanation via a separately trained network, which maps an explanation to a label.
We show that our framework is able to make \attention\footnote{We use the pretrained model from \url{https://github.com/OanaMariaCamburu/e-SNLI}.} generate a significant number of inconsistent explanations.
We highlight that our final goal is not a label attack, even if, for this particular model in which the label is predicted solely from the explanation, we implicitly also have a label attack with high probability.\footnote{Their Explanation-to-Label component had $96.83\%$ test accuracy.}
In our experiments, we set $\bfxd$ as the premise (as this represents the given context in this task) and $\bfxi$ as the hypothesis. However, note that due to the nature of SNLI for which decisions are based mostly on commonsense knowledge, the explanations are most of the time independent of the premise, such as ``\emph{A dog is an animal.}'' --- hence, it would be possible to also reverse the premise and not just the hypothesis; we leave this as future work. 
For the \rvj~model, we use the same neural architecture and hyperparameters used by \citet{esnli} for \attention. \rvj~takes as input a premise-explanation pair, and produce a hypothesis.
Our trained \rvj~model is able to reconstruct \emph{exactly the same} (according to string matching) hypothesis with $32.78\%$ test accuracy. 

\paragraph{Creating $\mathcal{I}_{\bfe}$.}
To execute step (\ref{rules}), we employ negation and swapping explanations.
For negation, we simply remove the tokens ``not'' and ``n't'' if they are present. If these tokens appear more than once in an explanation, we create multiple inconsistencies by removing only one occurrence at a time. We do not attempt to add negation tokens, as this may result in grammatically incorrect sentences.
For swapping explanations, we note that the explanations in e-SNLI largely follow a set of label-specific templates. This is a natural consequence of the task and the SNLI dataset and not a requirement in the collection of the e-SNLI. For example, annotators often used ``\emph{One cannot X and Y simultaneously.}'' to explain a contradiction, ``\emph{Just because X, doesn't mean Y.}'' for neutral, or ``\emph{X implies Y.}'' for entailment.
Since any two labels are mutually exclusive, transforming an explanation from one template to a template of another label should automatically create an inconsistency.
For example, for the explanation of the contradiction ``\emph{One cannot eat and sleep simultaneously.}'', we match \emph{X} to ``\emph{eat}'' and \emph{Y} to ``\emph{sleep}'', and create the inconsistent explanation ``\emph{Eat implies sleep.}'' using the entailment template ``\emph{X implies Y.}''. Thus, for each label, we created a list of the most used templates that we manually identified among e-SNLI, which can be found in \cref{assec:templates}. A running example of creating inconsistent explanations by swapping is given in \cref{run-ex}. 
%

%
%
If there is no negation and no template match, we discarded the instance. In our experiments, we only discarded $2.6\%$ of the SNLI test set.
%
%

%
We note that this procedure may result in grammatically or semantically incorrect inconsistent explanations. However, as we will see below, our \rvj~performed well in generating correct and relevant reverse hypotheses even when its input explanations were not correct. This is not surprising, because \rvj~has been trained to output ground-truth hypotheses. 

The rest of the steps follow as described in (\ref{b}) - (\ref{check_rules}). 

\paragraph{Results and discussion.}
We identified a total of $1044$ pairs of inconsistent explanations starting from the SNLI test set, which contains $9824$ instances. First, we noticed that there are, on average, $1.93 \pm 1.77$ distinct reverse hypotheses giving rise to a pair of inconsistent explanation. Since the hypotheses are distinct, each of these instances is a separate valid adversarial inputs. However, if one is strictly interested in the number of distinct pairs of inconsistent explanations, then, after eliminating duplications, we obtain $540$ pairs of such inconsistencies.

Secondly, since the generation of natural language is always best evaluated by humans, we manually annotated $100$ random distinct pairs. We found that $82\%$ of the reverse hypotheses form realistic instances together with the premise. We also found that the majority of the unrealistic instances are due to a repetition of a token in the hypothesis. For example, ``\emph{A kid is riding a helmet with a helmet on training.}'' is a generated reverse hypothesis which is just one token away from a perfectly valid hypothesis. 

Given our estimation of $82\%$ to be inconsistencies caused by realistic reverse hypotheses, we obtained a total of $\sim\!\!443$ distinct pairs of inconsistent explanations. While this means that our procedure only has a success rate of $\sim\!\!4.51\%$, it is nonetheless alarming that this very simple and under-optimized adversarial framework detects a significant number of inconsistencies on a model trained on $\sim\!\!570$K examples. In \cref{inconsist-examples}, we see three examples of detected inconsistencies. More examples can be found in \cref{more-examples}.

\paragraph{Manual scanning.} We were curious to what extent one can find inconsistencies via a brute-force manual scanning. We performed three such experiments, with no success. On the contrary, we noticed a good level of robustness against inconsistencies when scanning through the generic adversarial hypotheses introduced by \citet{behaviour}. The details are in \cref{man-scan}. 


\section{Related Work}\label{related}
%

An increasing amount of work focuses on providing natural language, free-form explanations \citep{esnli,cars,zeynep,DBLP:conf/eccv/HendricksARDSD16} as a 
more comprehensive and user-friendly alternative to other forms of explainability, such as feature-based explanations~\citep{lime,shap}.
%
In this work, we bring awareness to the risk of generating inconsistent explanations.
Similarly, \citet{grounding} identify the risk of mentioning attributes from a strong class prior without any evidence being present in the input. 
%
%
%
%
%
%
\paragraph{Generating adversarial examples.}
Generating adversarial examples is an active research area in natural language processing~\citep{asurvey,DBLP:journals/corr/abs-1902-07285}.
%
%
%
However, most works build on the requirement that the adversarial input should be a small perturbation of an original input~\citep{DBLP:journals/corr/abs-1711-02173,DBLP:conf/cvpr/HosseiniXP17,keywords}, or should be preserving the semantics of the original input~\citep{DBLP:journals/corr/abs-1804-06059}. 
Our setup does not have this requirement, and any pair of task-realistic inputs that causes the model to produce inconsistent explanations suffices.
%
%
%
%
%
Most importantly, to our knowledge, no previous adversarial attack for sequence-to-sequence models generates \emph{full} target sequences. 
For instance, \citet{keywords} 
require the presence of pre-defined tokens anywhere in the target sequence: they only test with up to $3$ required tokens, and their success rate dramatically drops from $99\%$ for 1 token to $37\%$ for 3 tokens for the task of summarization.
Similarly, \citet{DBLP:conf/iclr/ZhaoDS18} proposed an adversarial framework for 
adding and removing tokens in the target sequence for the task of machine translation.
Our scenario would require as many tokens as the desired adversarial explanation, and we also additionally need them to be in a given order, thus tackling a much challenging task. 
Finally, \citet{DBLP:conf/conll/Minervini018} attempted to find inputs where a model trained on SNLI violates a set of logical constraints. 
However, their method needs to enumerate and evaluate a potentially very large set of perturbations of the inputs. 
Besides the computational overhead, it also may easily generating ungrammatical inputs. Moreover, their scenario does not address the question of automatically producing undesired (inconsistent) sequences.
%
%

%


%


%
\section{Summary and Outlook}
We drew attention that models generating natural language explanations are prone to producing inconsistent explanations. 
This concern is general and can have a large practical impact. For example, users would likely not accept a self-driving car if its explanation module is prone to state that ``\emph{The car accelerates because there are people crossing the intersection.}''.
We introduced a generic framework for identifying such inconsistencies and showed that the best existing model on e-SNLI can generate a significant number of inconsistencies.
Future work will focus on developing more advanced procedures for detecting inconsistencies, and on building robust models that do not generate inconsistencies.
\paragraph{Acknowledgments.}
This work was supported by a JP Morgan PhD Fellowship, the Alan Turing Institute under the EPSRC grant EP/N510129/1, the EPSRC grant EP/R013667/1, the AXA Research Fund, and the EU Horizon 2020 Research and Innovation Programme under the grant 875160.

\bibliography{bibliography}
\bibliographystyle{acl_natbib}

\clearpage
\newpage
\appendix

\section{e-SNLI Explanations Templates}~\label{assec:templates}

Below we present the list of templates that we manually found to match most of the e-SNLI explanations \citep{esnli}. We recall that during the collection of the dataset \citet{esnli} did not impose any template, they were a natural consequence of the task and SNLI dataset.

Here, ``\emph{subphrase1}/\emph{subphrase2}/...'' means that a separate template is to be considered for each of the subphrases. X and Y are the key elements that we want to identify and use in the other templates in order to create inconsistencies. ``[...]'' is a placeholder for any string, and its value is not relevant. Subphrases placed between round parenthesis (for example, ``(the)'' or ``(if)'') are optional, and two distinct templates are formed one with and one without that subphrase.

\paragraph{Entailment Templates}

\begin{itemize}
\item \ X is/are a type of Y
\item \ X implies Y
\item \ X is/are the same as Y
\item \ X is a rephrasing of Y 
\item \ X is a another form of Y
\item \ X is synonymous with Y
\item \ X and Y are synonyms/synonymous
\item \ X and Y is/are the same thing
\item \ (if) X , then Y
\item \ X so Y
\item \ X must be Y
\item \ X has/have to be Y
\item \ X is/are Y
\end{itemize}

\paragraph{Neutral Templates}

\begin{itemize}
\item \ not all    X are Y
\item \ not every    X is Y
\item \ just because    X does not/n't mean/imply Y
\item \   X is/are not necessarily Y
\item \   X does not/n't have to be Y
\item \   X does not/n't imply/mean Y

\end{itemize}

\paragraph{Contradiction Templates}

\begin{itemize}
\item \ ([...]) cannot/can not/ca n't (be) X and Y at the same time/simultaneously 
\item \ ([...]) cannot/can not/ca n't (be) X and at the same time Y
\item \   X is/are not (the) same as Y
\item \ ([...]) is/are either X or Y
\item \   X is/are not Y
\item \   X is/are the opposite of Y
\item \ ([...]) cannot/can not/ca n't (be) X if (is/are)~Y
\item \   X is/are different than Y
\item \   X and Y are different ([...])
\end{itemize}

\subsection{Running Example for Creating Inconsistencies by Swapping between Templates of Explanations}
\label{run-ex}


Consider the explanation $\bfe=$``\textit{Dog is a type of animal.}'' which may arise from a model explaining the instance $\bfx=$ (premise: ``\textit{A dog is in the park.}'', hypothesis: ``\textit{An animal is in the park.}''). We identify that $\bfe$ matches the template ``X is/are a type of Y'' with X = ``dog'' (we convert to lowercase) and Y = ``animal''. We generate the list $\mathcal{I}_{\bfe}$ by replacing X and Y in each of the neutral and contradictory templates listed above with the exception of those that contain ``[...]'' in order to avoid guessing the placeholder. We obtain $\mathcal{I}_{\bfe}$ as:

\begin{itemize}
\item \ not all dog are animal
\item \ not every dog is animal
\item \ just because dog does not/n't mean/imply animal
\item \   dog is/are not necessarily animal
\item \   dog does not/n't have to be animal
\item \   dog does not/n't imply/mean animal

\item \ cannot/can not/ca n't (be) dog and animal at the same time/simultaneously 
\item \ cannot/can not/ca n't (be) dog and at the same time animal
\item \   dog is/are not (the) same as animal
\item \ is/are either dog or animal
\item \   dog is/are not animal
\item \   dog is/are the opposite of animal
\item \  cannot/can not/ca n't (be) dog if (is/are) animal
\item \   dog is/are different than animal
\item \   dog and animal are different
\end{itemize}




\begin{table*}[t]
\resizebox{\textwidth}{!}{%
\begin{tabular}{p{12cm}|p{12cm}}

\toprule

\multicolumn{2}{c}{\begin{tabular}[c]{@{}c@{}} \textsc{Premise:} Biker riding through the forest.\end{tabular}}

\\

\begin{tabular}[c]{@{}l@{}}
\textsc{Original Hypothesis:} Man riding motorcycle on highway. \\
\textsc{Predicted Label:} contradiction \\
\textsc{Original explanation:} {\bf Biker and man are different.}\end{tabular} &

\begin{tabular}[c]{@{}l@{}}
\textsc{Reverse Hypothesis:} A man rides his bike through the forest. \\
\textsc{Predicted label:} entailment \\
\textsc{Reverse explanation:} {\bf A biker is a man.} \end{tabular}

\\

\midrule

\multicolumn{2}{c}{\begin{tabular}[c]{@{}c@{}} \textsc{Premise:} A hockey player in helmet.\end{tabular}} 

\\

\begin{tabular}[c]{@{}l@{}}
\textsc{Original Hypothesis:} They are playing hockey \\
\textsc{Predicted Label:} entailment \\
\textsc{Original explanation:} {\bf A hockey player in helmet is playing hockey.} \end{tabular}
& 
\begin{tabular}[c]{@{}l@{}}
\textsc{Reverse Hypothesis:} A man is playing hockey. \\
\textsc{Predicted Label:} neutral \\
\textsc{Reverse explanation:} {\bf A hockey player in helmet doesn't imply playing} \\ \bf{hockey.} \end{tabular} 

\\

\midrule

\multicolumn{2}{c}{\begin{tabular}[c]{@{}c@{}}
\textsc{Premise:} A blond woman speaks with a group of young dark-haired female students carrying pieces of paper. \end{tabular}}

\\

\begin{tabular}[c]{@{}l@{}}
\textsc{Original Hypothesis:} A blond speaks with a group of young \\ dark-haired woman students carrying pieces of paper. \\
\textsc{Predicted label:} entailment \\
\textsc{Original explanation:} \textbf{A woman is a female.}\\
\end{tabular}                                
& 
\begin{tabular}[c]{@{}l@{}}
\textsc{Reverse Hypothesis:}The students are all female. \\
\textsc{Predicted label:} neutral \\
\textsc{Reverse explanation:} \textbf{The woman is not necessarily female.} \end{tabular}

\\
\midrule

\multicolumn{2}{c}{\begin{tabular}[c]{@{}c@{}}
\textsc{Premise:} The sun breaks through the trees as a child rides a swing. \end{tabular}}

\\

\begin{tabular}[c]{@{}l@{}}
\textsc{Original Hypothesis:} A child rides a swing in the daytime. \\
\textsc{Predicted label:} entailment \\
\textsc{Original explanation:} \textbf{The sun is in the daytime.} \\
\end{tabular}                                
& 
\begin{tabular}[c]{@{}l@{}}
\textsc{Reverse Hypothesis:} The sun is in the daytime. \\
\textsc{Predicted label:} neutral \\
\textsc{Reverse explanation:} \textbf{The sun is not necessarily in the daytime.} \end{tabular}

\\

\midrule

\multicolumn{2}{c}{\begin{tabular}[c]{@{}c@{}}
\textsc{Premise:} A family walking with a soldier. \end{tabular}}

\\

\begin{tabular}[c]{@{}l@{}}
\textsc{Original Hypothesis:} A group of people strolling. \\
\textsc{Predicted label:} entailment \\
\textsc{Original explanation:} \textbf{A family is a group of people.} \\
\end{tabular}                                
& 
\begin{tabular}[c]{@{}l@{}}
\textsc{Reverse Hypothesis:} A group of people walking down a street. \\
\textsc{Predicted label:} contradiction \\
\textsc{Reverse explanation:} \textbf{A family is not a group of people.} \end{tabular}

\\

\bottomrule

\end{tabular}%
}
\caption{More examples of inconsistent explanations detected with our method.}
\label{more-ex}
\end{table*}

\section{More Examples of Detected Inconsistencies}
\label{more-examples}
In \cref{more-ex}, we provide more examples of inconsistent explanations detected with our method.

\section{Manual Scanning}
\label{man-scan}
We performed three experiments of manually scanning. First, we manually analyzed the first $50$ instances in the test set without finding any inconsistency. However, these examples were involving different concepts, thus decreasing the likelihood of finding inconsistencies. To account for this, in our second experiment, we constructed three groups around the concepts of \emph{woman}, \emph{prisoner}, and \emph{snowboarding}, by simply selecting the explanations in the test set containing these words. We selected these concepts,  because our framework detected inconsistencies about them --- examples are listed in \cref{inconsist-examples} and \cref{more-ex}.

For \emph{woman}, we obtained $1150$ examples in the test set, and we looked at a random sample of $20$, among which we did not find any inconsistency.
For \emph{snowboarding}, we found $16$ examples in the test set and again no inconsistency among them.
For \emph{prisoner}, we only found one instance in the test set, so we had no way to find out that the model is inconsistent with respect to this concept simply by scanning the test set.

We only looked at the test set for a fair comparison with our method that was only applied on this set.

However, we highlight that, even if the manual scanning would have been successful, it should not be regarded as a proper baseline, since it does not bring the same benefits as our framework. Indeed, manual scanning requires considerable human effort to look over a large set of explanations in order to find if any two are inconsistent. Even a group of only $50$ explanations required us a non-negligible amount of time. Moreover, restricting ourselves to the instances in the original dataset would clearly be less effective than being able to generate new instances from the dataset's distribution. Our framework addresses these issues and directly provides pairs of inconsistent explanations. Nonetheless, we considered this experiment useful for illustrating that the explanation module does not provide inconsistent explanations in a frequent manner. 

In our third experiment of manual scanning, we experimented with a few manually created hypotheses from \citet{behaviour}, which had been shown to induce confusion at the label level. We were pleased to notice a good level of robustness against inconsistencies. For example, for the neutral pair (premise: ``\emph{A bird is above water.}'', hypothesis: ``\emph{A swan is above water.}''), we get the explanation ``\emph{\textbf{Not all birds are a swan.}}'', while when interchanging bird with swan, i.e., for the pair (premise: ``\emph{A swan is above water.}'', hypothesis: ``\emph{A bird is above water.}''), \attention~generates the explanation ``\emph{\textbf{A swan is a bird.}}'', showing a good understanding of the relationship between the entities ``swan'' and ``bird''.
Similarly, interchanging ``child'' with ``toddler'' in (premise: ``\emph{A small child watches the outside world through a window.}'', hypothesis: ``\emph{A small toddler watches the outside world through a window.}'') does not confuse the model, which outputs ``\textbf{\emph{Not every child is a toddler.}}'' and ``\emph{\textbf{A toddler is a small child.}}'', respectively. Further investigation on whether the model can be tricked on concepts where it seems to exhibit robustness, such as \emph{toddler} or \emph{swan}, are left for future work.

\end{document}